%% file: main.tex
\documentclass{article}

\usepackage{microtype}
\usepackage{graphicx}
\usepackage{subfigure}
\usepackage{booktabs} 
\usepackage{amsmath} 
\usepackage{amssymb}
\usepackage{color}
\usepackage[ruled,vlined]{algorithm2e}

\usepackage{hyperref}

\usepackage[accepted]{icml2020}

\icmltitlerunning{Deep Reinforcement Learning with Linear Quadratic Regulator Regions}

\begin{document}

\twocolumn[
\icmltitle{Deep Reinforcement Learning with Linear Quadratic Regulator Regions}



\icmlsetsymbol{equal}{*}

\begin{icmlauthorlist}
\icmlauthor{Gabriel I.~Fernandez}{romela}
\icmlauthor{Colin Togashi}{romela}
\icmlauthor{Dennis W.~Hong}{romela}
\icmlauthor{Lin F.~Yang}{ee}
\end{icmlauthorlist}

\icmlaffiliation{romela}{Robotics \& Mechanisms Laboratory, Mechanical \& Aerospace Engineering Department, University of California Los Angeles, California, USA}
\icmlaffiliation{ee}{Department of Electrical \& Computer Engineering, University of California Los Angeles, California, USA. \href{https://github.com/gabriel80808/pyrol}{Algorithm Implementation}}

\icmlcorrespondingauthor{Gabriel I.~Fernandez}{gabriel808@g.ucla.edu}

\icmlkeywords{Robotics, Legged, Robots, Reinforcement Learning, Deep, RL, LQR, Linear Quadratic Regulator, Machine Learning, Simulation to Real, Sim-to-Real, pendulum, ICML}

\vskip 0.3in
]



\printAffiliationsAndNotice{}  
\begin{abstract}
Practitioners often rely on compute-intensive domain randomization to ensure reinforcement learning policies trained in simulation can robustly transfer to the real world. Due to unmodeled nonlinearities in the real system, however, even such simulated policies can still fail to perform stably enough to acquire experience in real environments. In this paper we propose a novel method that guarantees a stable region of attraction for the output of a policy trained in simulation, even for highly nonlinear systems. Our core technique is to use ``bias-shifted'' neural networks for constructing the controller and training the network in the simulator. The modified neural networks not only capture the nonlinearities of the system but also provably preserve linearity in a certain region of the state space and thus can be tuned to resemble a \emph{linear quadratic regulator} that is known to be stable for the real system. We have tested our new method by transferring simulated policies for a swing-up inverted pendulum to real systems and demonstrated its efficacy.
\end{abstract}
\input{Sections/1Introduction}
\input{Sections/2Related_Works}
\input{Sections/3Preliminaries}
\input{Sections/4Shifted_Bias}
\input{Sections/5Framework}
\input{Sections/6Results}
\input{Sections/7Conclusion}

\clearpage
\bibliography{bibliography}
\bibliographystyle{icml2020}
\clearpage
\appendix
\input{Sections/A1TD3Algorithm}
\input{Sections/A2AllTrainedGraphs}
\input{Sections/A3PolicyEvolution}
\end{document}

%% file: Sections/1Introduction.tex
\section{Introduction}
Robotic systems have largely depended on classical control theory for its theoretical performance guarantees. As the demand increases for robots to accomplish ever more complex tasks, these traditional methods cease to satisfy expectations. Deep reinforcement learning models have achieved a high level of performance \cite{duan2016benchmarking} on aforementioned, nontrivial tasks but have been predominantly confined to simulated environments. Often times the unmodeled nonlinearities of the hardware render simulation-trained policies useless or even worse endanger the system's operability.

To overcome these issues. practitioners have become increasingly dependent on exponentially more data samples with an approach called domain randomization (DR). This method trains the policy on a range of simulation parameters, usually including the hardware parameters as a subset. Researchers have been successsful in applying this technique on drones and robotic hands to complete difficult tasks \cite{tobin2017domain, loquercio2019deep,  akkaya2019solving}. Unfortunately, such a method usually requires expensive computation, e.g., \cite{akkaya2019solving} used 64 NVIDIA V100 GPUs and 920 worker machines with 32 CPU cores for their Rubik's cube task, and the total amount of training experience in simulation for all their experiments amounted to approximately 13 thousand years. Another direction builds better simulated models by capturing the nonlinearities of the actual system through deep learning, e.g., \cite{hwangbo2019learning}. Building such models are challenging, often times requiring meticulous handcrafting for specific systems. Unfortunately, neither DR nor learned simulators have guarantees for convergence or behavior.

In this paper, we seek for a more ideal approach that leverages the advantages from linear control theory and deep reinforcement learning to efficiently train policies in simulation yet be able to run on hardware with guarantees. Specifically, this policy class would capture both the linear region and nonlinear region of a system, where in the linear region it behaves like a classical controller and in the nonlinear region, a learned policy. Then, when transferring the policy onto hardware, the linear region can be used to fine tune the parameters so that the policy in the nonlinear region begins to perform stably. 

Fortunately, for many practical systems, e.g., robots, drones and pendulums, pre-existing robust linear controllers perform well under linearization assumptions, which albeit cannot outperform the best trained deep-learning based policies. These controllers nevertheless inspire us to ask the following question: \emph{Can we embed prior-known linear feedback controllers into a neural network architecture such that the resulting policy can be as rich as the standard neural network and as stable as a linear controller?} In this paper, we answer this question affirmatively by proposing a novel neural network structure called ``bias-shifted'' neural networks to construct the policy.

\begin{figure}[!h]
	\centering
		\includegraphics[scale=0.13]{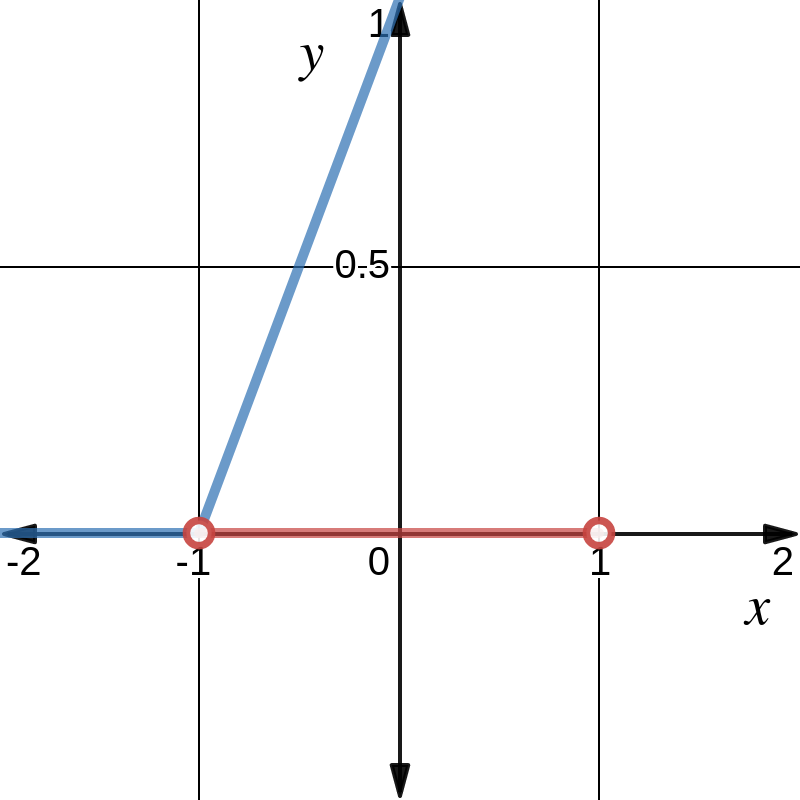}
	~~
		\includegraphics[scale=0.13]{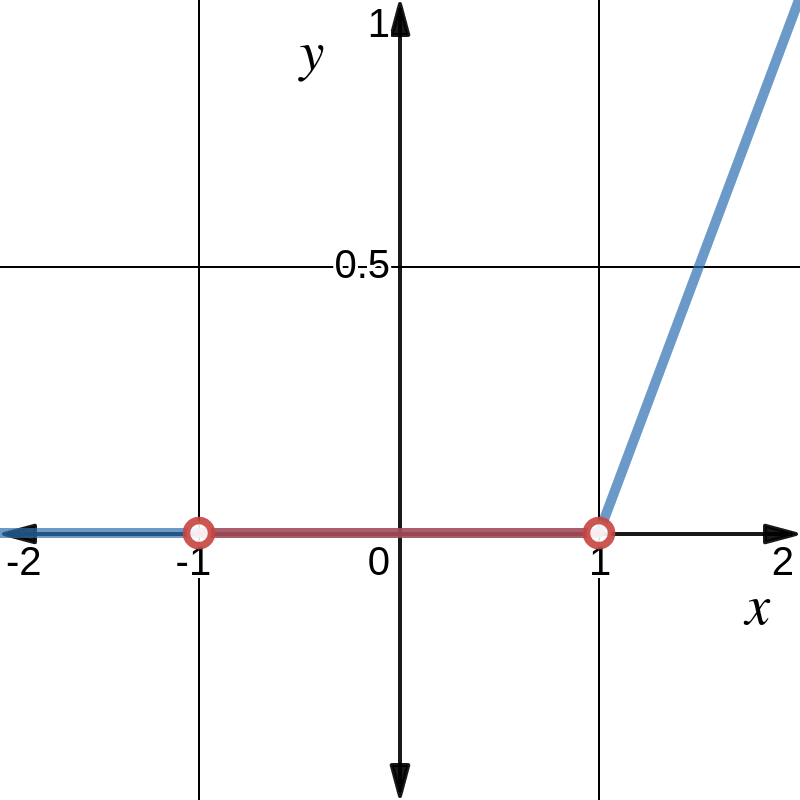}
	\caption{\emph{Left}: The blue line depicts a ReLU activation function with a bias of $+1$, creating two different linear regions on both sides of $x=-1$. \emph{Right}: The blue line again shows ReLU but with a bias of $-1$. Notice that in both graphs the domain highlighted in red only contains affine, $y=x+b$, (\emph{Left}) or linear, $y=0$, (\emph{Right}) portions of each function.
	}
	\label{Fig:bias_constraints}
\end{figure}

The core construction comes from the following observation: The nonlinearity of a standard feedforward network comes from its activation functions, which usually can be decomposed into (approximate) linear parts\footnote{
``Linear'' stands for affine or linear throughout for simplicity and consistency with \cite{montufar2014number, pascanu2013number}.
}. For instance, ReLU, defined as $\sigma(x):=\max(x, 0)$, has two linear parts on the intervals $(\infty, 0)$ and $[0, \infty)$ (see Fig.~\ref{Fig:bias_constraints}).  Here $x=0$ defines a transition point for ReLU. Since an input region, $[-m, m]$, contains the point, zero, the overall response of ReLU over this region is no longer linear. But, if we add a bias to function, e.g., change ReLU to $\sigma'(x):=\max(x + b, 0)$ where $|b|\ge m$, and then $\sigma'(x)$ performs linearly in the region $[-m, m]$. We leverage this observation by explicitly ``bias-shifting'' a neural network before and during training to preserve linearity for a desired range of inputs, i.e., restrict the bias of every layer to at least have certain magnitude. As we are still using the same set of activation functions and number of hidden layers, the nonlinearity of the system outside of that range can still be preserved. 

We apply our ``bias-shifted'' neural network  with an Actor-Critic framework \cite{fujimoto2018addressing} for system control and modify the actor network to comply with our ``bias-shifted'' structure to maintain a linear region. Within this linear region, we fit the network with a pre-existing linear controller, e.g., a linear optimal controller: \emph{linear quadratic regulator} (LQR), by either adding a regularizer or modifying the weights of the last layer of the network, or both. We experiment with this new method in controlling a simple nonlinear system: the swing-up pendulum. Despite modeling discrepancies, the new controller with the bias-shifted network was effective and robust in the linear region.

%% file: Sections/2Related_Works.tex
\section{Related Works}
We are not aware of any literature involving the creation of an explicit linear region in a feedforward network. Although, the notion of linear regions in deep learning neural networks has been used as a metric to theoretically determine expressivity of a particular architecture \cite{montufar2014number, pascanu2013number}.

To our knowledge there has not been any work done on directly adjusting the weights of a network to output an LQR controller for a region of the state space. Rather, a number of publications have taken advantage of LQR's structure for efficiency purposes. \cite{du2019continuous} builds a decoder that maps observations to a linear representation where LQR can be used. \cite{marco2017design} uses a Gaussian process with kernels of an LQR structure. \cite{bradtke1993reinforcement} uses the LQR structure for the $Q$-value function. Incorporating LQR in these manners can come at the cost of limiting expressivity power of the actual controller itself.

Recently, researchers have had some success transferring learned policies on to hardware and completed complex tasks using DR \cite{akkaya2019solving, peng2018sim} and DR with actuator modelling in simulation \cite{hwangbo2019learning}. There has been a wave of publications applying DR and incorporating real data to improve the simulators accuracy \cite{tan2018sim, chebotar2019closing, james2019sim, golemo2018sim, carlson2019sensor}. Incorporating real knowledge into the simulator is always beneficial but requires nontrivial effort. Our method, on the other hand, directly incorporates knowledge about the system into the policy itself.

Another approach along the lines of DR is to simply develop robust policies \cite{turchetta2019robust, han2019h_, muratore2019assessing}. Bias-shifted networks can be used to output robust policies in the linear region if desired as well. Additionally, robustness metrics can be added 
to train our policy, hence giving a robust policy in the nonlinear region.

%% file: Sections/3Preliminaries.tex
\section{Preliminaries}
This section covers fundamental concepts used throughout.

\subsection{Notations}
Throughout the paper, capital letters are used for matrices, e.g., $W, A, B, Q, R$, and lower-case letters for vectors, e.g., $x, u, v, b$. The dimensions of the vectors and matrices are adjusted according to context unless otherwise specified. Greek letters denote functions, e.g., $\sigma:\mathbb{R}\rightarrow \mathbb{R}$. A function applied to a vector, e.g., $\sigma(x)$, represents component-wise operation. Applying arithmetic operations to vectors also indicates component-wise operation, e.g., $\in, +, -, =, |\cdot|$. For a vector $b$, the $j$-th element is taken as $b_j$. Lastly, $0$ stands for a vector with all zeros.

\subsection{Deep Neural Network}
Deep neural networks have become the foundation for many algorithms due to their expressiveness. One of the most common choices is the feedforward, fully-connected network. This class of neural networks consists of sequentially connected layers made up of linear combinations of the input followed by nonlinear activation functions, $\sigma$. The linear combinations are weighted, $W$, and have a bias, $b$. The output, $f(x)$, of a single layer expressed in terms of its input, $x$, for a single layer is given as follows:
\begin{equation*}
    f(x)=\sigma(W_1x + b_1)
\end{equation*}
And when multiple layers are sequentially connected:
\begin{equation*}
    f(x)=\sigma(W_n \ \sigma_{n-1}(... \ \sigma_1(W_1x+b_1) + ...) + b_n).
\end{equation*}

\subsection{Linear Quadratic Regulator}
\label{Sec:Linear Quadratic Regulator}
\begin{figure}[!h]
    {\centering
		\includegraphics[scale=0.25]{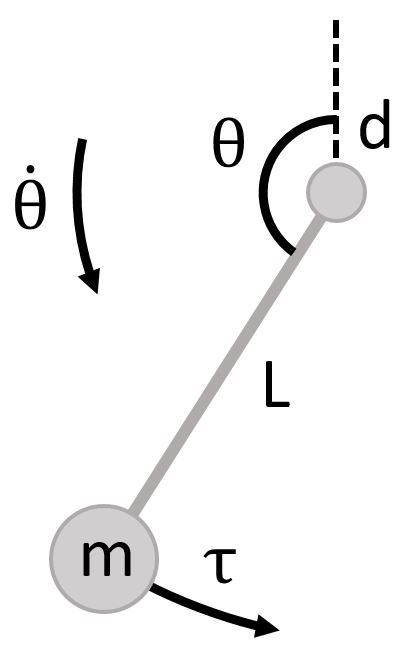}
    ~
		\includegraphics[scale=0.085]{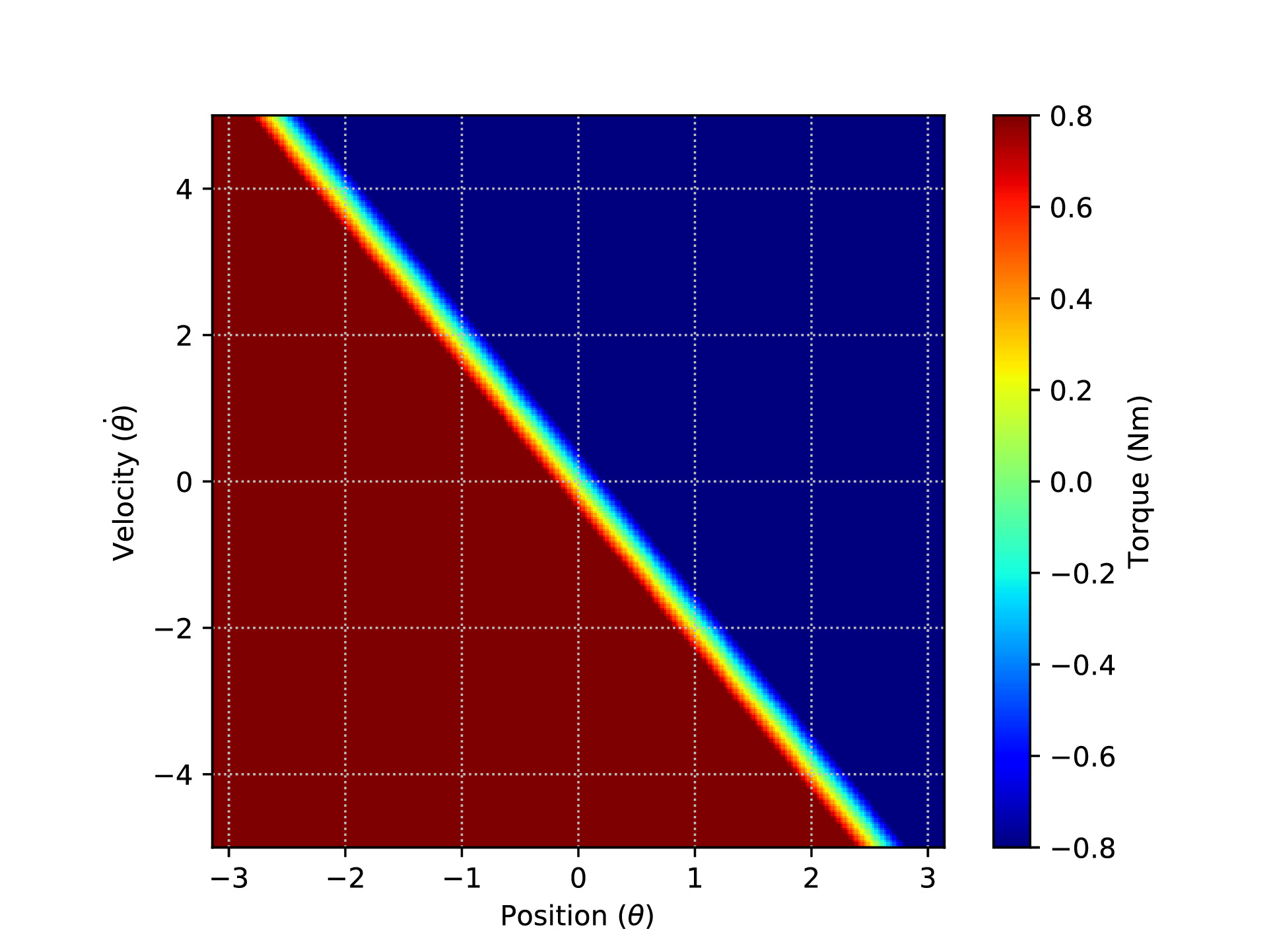}
	}
	\caption{\label{fig:pendulum}
	\emph{Left}: A pendulum system with mass $m$, length $L$, damping $d$, position $\theta$, velocity $\dot{\theta}$, and torque $\tau$. Coordinates are defined such that a counterclockwise direction denotes a positive position and torque. 
	\emph{Right}: The LQR controller for the pendulum environment sampled at different points of the state space.
	}
	\label{Fig:pendulum}
\end{figure}

The dynamics of a linear time invariant, continuous time system with state transition matrix, $A$, state, $x$, control input, $u$, and control matrix, $B$, can be formulated as:
\begin{equation}
    \dot{x}(t) = Ax(t) + Bu(t), \qquad x(0) = x_0.
\label{Eq:linear_sys}
\end{equation}
For LQR, the optimal cost function is defined to have a quadratic relationship with respect to $x$ and $u$ with weight matrices $Q \succeq 0$ and $R \succ 0$, respectively:
\begin{equation}
    J = \int^\infty_0 x(t)^TQx(t) + u(t)^TRu(t)dt
\label{Eq:lqr_cost_int}
\end{equation}
Assuming the matrix pair $(A, B)$ from \eqref{Eq:linear_sys} forms a controllable system \cite{hespanha2018linear}, the LQR optimal controller can be formulated using a linear state feedback gain, $K$:
\begin{equation}\textstyle
    u(t) = -Kx(t)
\label{Eq:lqr_control}
\end{equation}
where $K = -R^{-1}B^TP$, and $P$ is the solution to the continuous algebraic Ricatti equation given by:
\begin{equation*}
    A^{\top} P + PA -PBR^{-1}B^{T}P + Q = 0.
\end{equation*}
The optimal cost, the lowest amount of cost to achieve the goal state, simplifies to $J = x_0^TPx_0$ where $x_0$ is the initial state. The LQR solution yields a closed form expression for a linear controller and a quadratic optimal cost. An LQR controller on a pendulum system is depicted in Fig.~\ref{fig:pendulum}.

\subsection{Reinforcement Learning Actor-Critic Algorithms}
In reinforcement learning (RL), the agent aims to minimize the cost of controlling a system without any prior knowledge of $A$ and $B$. The controller, $u$, is defined as a policy, $\pi$, i.e.,
$ \pi(x(t))=u(t)$. The goal is to find an optimal policy such that the 
$Q$-function minimizes costs and maximizes rewards. The $Q$-function, corresponding to the cost function $J$ in LQR, is given as the following\footnote{The continuous form of the Q-value is an integral. But in practice, the discretized approximation is usually used.}:
\begin{equation*}
    Q(x(t), u(t))=\mathbb{E}\bigg[\sum^N_{t'=t}\gamma^{t'}r_{t'+1}\big|x_0=x(t), u_0=u(t)\bigg]
\end{equation*}

There are many algorithms proposed for RL \cite{li2017deep}. In this paper, we use the actor-critic algorithm, Twin Delayed Deep Deterministic policy gradient (TD3) from \cite{fujimoto2018addressing}, to control our systems. We choose this algorithm since its benchmarks outperform its peers, and its simplistic design makes it straightforward to adjust the architecture for our purposes.
We briefly introduce the algorithm here. There are two types of neural networks: the actor network, $\pi_{\phi}$, and the critic network $Q_{\theta}$. The actor network, $\pi_{\phi}$, represents the policy to be learned, and the critic network, $Q_{\theta}$, represents the $Q$-function. These networks are randomly initialized. During each epoch of the training, the agent uses $\pi_{\phi}$ (with noise added for exploration) to control the system and collect data, where $\phi$ denotes the parameters of the network $(W, b)_{\phi}$. Once a transition (a state-action-reward-state tuple representing an action executed at a state and the observed reward and new state) is collected, we update $Q_{\theta}$  by performing a dynamic programming step, where $\theta$ denotes the parameters of the network $(W, b)_{\theta}$. After a sufficient amount of transitions have been collected, we update the policy $\pi_{\phi}$ by following its gradient, which is computed using $Q_{\theta}$. The full algorithm is presented in Algorithm~\ref{Alg:TD3Original} in Appendix~\ref{Sec:original_td3}. Note that in the actual implementation, there are two networks $Q_{\theta_1}$ and $Q_{\theta_2}$, both representing critics. These subtleties are designed to specifically address overestimation bias in the $Q$-function \cite{fujimoto2018addressing}.

%% file: Sections/4Shifted_Bias.tex
\section{Shifted-Bias Neural Network}
\label{Sec: Shifted-Bias Neural Network}
In this section we explain how to ``bias-shift'' a network to preserve the linear region. In principle, any activation function can be used in our approach as long as it either contains a linear or approximately linear segment. Throughout this paper, ReLU\footnote{One could argue that $sigmoid$ and similar activations provide an linear region to begin with if everything stays close to zero. This would require all the biases to be set to zero, which can be considered a ``bias-shift''. During training, $sigmoid$-like functions would still require the other two ``bias-shifted'' approaches: maintenance of the linear region during training and tuning post-training. Moreover, ReLU outperforms these types of activations which experience saturation, resulting in vanishing gradients and decreased learning rates \cite{chang2015batch, dahl2013improving}.} will be used to demonstrate and formulate the algorithms. The generalized notation of $\sigma$ for activation functions will be kept throughout to indicate this can be used with other activation types. Each type of activation unit will require its own slight modifications to the constructions presented here. As explained in the introduction, shifting the bias of a single ReLU is straightforward. However, with multi-layer, deep networks, it is still not clear how to create a linear region. Fortunately, after expanding a deep network layer by layer, a clear formulation emerges.

\subsection{Linear Region of Neural Networks}
\label{Sec:Linear Region of Neural Networks}
\paragraph{Linear Region of a One Layer Network}
Let us first consider a single layer network:
\[
f(x) =  \sigma(W_1 x + b_1),
\]
where $b_1$ is a $d$-dimensional vector. We first shift $b_1$ such that each of its entries has magnitude greater than or equal to $m>0$. As shown in the introduction, the linear region of the input is given by:
\[
\mathcal{L}_1:= \{x:  (W_1 x)_j\in [-m, m], \forall j\in [d]\}.
\]

\paragraph{Linear Region of a Two Layer Network}
Let us now consider a two-layer network:
\[
f(x) = \sigma(W_2 \sigma(W_1 x + b_1) + b_2), 
\]
Again, shifting $b_1$ such that each of its entries has magnitude greater than or equal to $m$, we immediately get:
\[
\forall x\in \mathcal{L}_1, j\in[d]:~ \sigma(W_1 x + b_1)_j = \max((W_1x)_j+b_{1,j}, 0)
\]
Note that $0$ is a linear function. Rewriting it as:  
\[
\forall x\in \mathcal{L}_1:~ \sigma(W_1 x + b_1) =: W_1^{(m)} x + b_1^{(m)}
\]
where if $\sigma(W_1 x + b_1)_j =0$, then the $j$-th row of $W_1^{(m)}$ is $0$ and $b_{1,j}^{(m)} = 0$, otherwise they are equal to the $j$-th row of $W_1$ or $j$-th entry of $b_1$, respectively.
In other words, $W_1^{(m)}$ and $b_{1}^{(m)}$ denote the ``masked'' version for $W_1$ and $b_1$ in the linear region of the first layer, $\mathcal{L}_1$. With this notation, the two layer network can be rewritten as:
\[
f(x) = \sigma(W_2 W_1^{(m)} x + W_2 b_1^{(m)} + b_2),
\]
which reduces to a single layer ReLU function. Similarly, the linear region defined by the second layer is:
\[
\mathcal{L}_2:= \{x:  (W_2 W_1^{(m)} x)_j\in [-m_{2, j}, m_{2,j}], \forall j\in [d]\},
\]
where $m_{2,j}:=|(W_2 b_1^{(m)})_{j} + b_{2,j}|$, which is derived from the effective bias for the second layer, $\hat{b}_{2}$, that determines the activation at the ReLU function given as:
\[
\hat{b}_2 = W_2 b_1^{(m)} + b_{2}.
\]

\paragraph{Linear Region of a Multi-Layer Network}
We are now ready to define a ``bias-shifting'' framework to get a linear region for a full $n$-layer neural network.
For layer, $l$, in a ReLU activated network, the linear region becomes:
\[
\mathcal{L}_l:= \{x:  (W_l( W_{l-1} ...)^{(m)} x)_j\in [-m_{l,j}, m_{l,j}], \forall j\in [d]\}
\]
where $m_{l,j}:=|(W_l (W_{l-1}( ... )^{(m)} + b_{l-1})^{(m)})_j + b_{l,j}|$.

The effective bias at layer, $l$, that determines the activation at the ReLU function is:
\begin{equation}
\hat{b}_l = (W_l (W_{l-1}( ... )^{(m)} + b_{l-1})^{(m)}) + b_{l}.
\label{eq:effective_bias}
\end{equation}

Thus, at the last layer, $n$, the effective weight, $\hat{W}_{n}$, and effective bias, $\hat{b}_{n}$, for the linear region are:
\begin{align}
    \hat{W}_n &= W_n( W_{n-1} ... )^{(m)}  = W_n\hat{W}_{n-1}^{(m)} \label{Eq:eff_weight} \\
    \hat{b}_n &= W_n (W_{n-1}( ... )^{(m)} + b_{n-1})^{(m)} + b_{n} \notag \\
    &= W_n\hat{b}_{n-1}^{(m)} + b_{n} \label{Eq:eff_bias}
\end{align}
Through defining the appropriate linear regions, the effective weight and the effective bias at the last layer can now be tuned through only changing the last layer weights, $W_n$, and the last layer biases, $b_n$, preserving the hidden layers.

%% file: Sections/5Framework.tex
\section{Algorithmic Framework}
\label{sec:alg}
In this section, we introduce our framework for modifying an actor-critic algorithm such that it enjoys the capabilities of an RL algorithm that encodes the nonlinearities of a system but additionally takes advantages of LQR for the stability. To begin, we assume that the system admits a controller (referred as linear controller throughout) that performs stably in a linear or approximately linear region of the hardware, which is the case as in our setting, e.g., a pendulum. We propose a ``bias-shifted'' network for training the actor network in TD3\footnote{Our framework can also be used to train the critic networks similarly. For presentation simplicity, we only describe the method for training the actor network.}. After training, the output actor will contain a linear region that can be used to fit existing linear controllers.

\paragraph{The Bias-shift Framework}
The ``bias-shift'' framework refers to three phases: initialization, training, and post-training processing. The initialization phase shifts the biases of the actor to be a desired magnitude corresponding to the input linear region. During training, regularization terms are added to the weight matrices to ensure the weights do not grow too large, which erodes the linearity provided by the bias. We also actively shift the biases in the actor network such that the magnitude never goes below the required linear region. To match the trained actor with the input linear controller, we apply a post-training processing stage. During this step, we adjust the last layer of the actor network so that its output matches that of the linear controller. To help counteract possible side effects in the post-training fitting, we also add a regularization term to the loss and use dropout to create feature redundancies.

\subsection{Initializing Linear Regions} \label{Sec:init_linear}
On the creation of a ``bias-shifted'' network, the effective bias of the $l$-th layer, $\hat{b}_{l}$, given by \eqref{eq:effective_bias} must be initialized such that $\hat{b}_{l} \ge m$ to form the linear region. The cumulative effects of $W_n\hat{b}_{n-1}$ as laid out in \eqref{Eq:eff_bias} need to be accounted for during initialization. In practice, we set biases in a sequential manner by uniformly sampling until the magnitude reaches the desired region size. As the combinations get more complex, we also can randomly sample biases and increment its value until the desired magnitude is met. Formally, initializing biases is given in Algorithm~\ref{Alg:bias_initialization}.

\begin{algorithm}[h]
   \caption{Bias Initialization}
   \label{Alg:bias_initialization}
\begin{algorithmic}[1]
    \STATE\textbf{Input: } linear region size: m$>$0;
    \STATE\textbf{Initialization:}
    \STATE Initialize biases: sample $b_l$ uniformly from\\
    $\;[(-c_bm,-m)\cup (m, c_bm)]$, using parameter $c_b>1$
    \IF{$l > 1$}
    \FOR{$\{b_{l,j}:W_l\hat{b}_{l-1}^{(m)} + b_l< m\}$ (Eq. \ref{Eq:eff_bias})}
    \WHILE{$|W_l\hat{b}_{l-1}^{(m)} + b_l| < m$}
    \STATE $b_{l,j} \leftarrow b_{l,j} + c_w\;b_{l,j}$ with parameter $c_w>0$
    \ENDWHILE
    \ENDFOR
    \ENDIF
\end{algorithmic}
\end{algorithm}

\subsection{Training of the Bias-Shifted Network}
Our detailed algorithm for training is presented in Algorithm~\ref{Alg:TD3training}. In the algorithm, we have two critic networks $Q_{\theta_1}$ and $Q_{\theta_2}$ to prevent overestimation bias \cite{fujimoto2018addressing}. The update of these networks follow the same way as in the original TD3 algorithm. We also have an actor network $\pi_{\phi}$, which we actively ``bias shift'' to avoid the collapsing of the linear region. The input parameters contain the linear region selector, $m$ on Line~18 of Algorithm~\ref{Alg:TD3training}. We actively shift all the bias in the actor with magnitudes less than $m$. To keep the magnitude of the weight matrices from increasing uncontrollably, we add a regularization term to Line~16. Note that these are the critical differences between our algorithm and the original TD3 framework.

\begin{algorithm*}[htb]
   \caption{Bias-Shifted TD3 Training}
   \label{Alg:TD3training}
\begin{algorithmic}[1]
    \STATE \textbf{Input:} linear region size: $m\ge 0$; number of iterations $T>0$; number of iterations before updating targets: $d>0$;
    \STATE \textbf{Initialization:}
    \STATE Initialize critic and actor networks: $Q_{\theta_1}$, $Q_{\theta_2}$ and $\pi_\phi$ and their target networks: $\theta_1' \leftarrow \theta_1$, $\theta_2' \leftarrow \theta_2$, $\phi'\leftarrow \phi$;
    \STATE Initialize replay buffer with random samples $\mathcal{B}$;
    \STATE 
    Initialize the biases using Algorithm~\ref{Alg:bias_initialization};
    \FOR{$t=1$ {\bfseries to} $T$}
    \STATE Let $a \gets \pi_\phi(s) + \epsilon$, where $\epsilon \sim \mathcal{N}(0, \sigma_e)$, where $\sigma_e$ is a parameter to be tuned;
    \STATE At state $s$, execute $a$ to obtain the next state $s'$, and a reward $r$; store ($s, a, r, s'$) in $\mathcal{B}$;
    \STATE Sample $N$ transitions $\{s_k, a_k, r_k, s_k'\}_{k=1}^{N}$ from $\mathcal{B}$
    \STATE For each $k\in[N]$: let
    $\tilde a_k \leftarrow \pi_\phi'(s'_k) + \epsilon$, where $\epsilon \sim \mathrm{clip}(\mathcal{N}(0, \sigma_s), -c, c)$ with tuning parameters $\sigma_s>0$ \& $c>0$ \\
    $\qquad\qquad\qquad\quad$ and $y_k \leftarrow r_k + \gamma  \min_{i=1,2} Q_{\theta_i'}(s_k', \tilde a_k)$, with discount factor $\gamma\in(0,1)$;
    \STATE Update critics using least square fitting: $\theta_i \gets \underset{\theta_i}{\text{argmin}}\: N^{-1}\sum_{k=1}^N(y_k-Q_{\theta_i}(s_k, a_k))^2$
    \IF{$t$ mod $d$}
    \STATE Add $\ell_2$ regularization (Sec. \ref{Sec:regularization}):
     $ J \gets J+ \mathrm{Reg}((W, b)_{\phi})$
    \STATE Update $\phi$ by policy gradient:
     $\nabla_\phi J(\phi) \gets \frac{1}{N}\sum\nabla_a Q_{\theta_1}(s, a)|_{a=\pi_\phi}\nabla_\phi\pi_\phi(s)+\nabla_\phi \mathrm{Reg}((W, b)_{\phi})$
    \STATE Ensure bias magnitude with shift:
    \textbf{if $|b_j| < m$ then}: $b_j \leftarrow m\;\text{sign}(b_j)$;
    \STATE Update target networks:$\theta_i' \leftarrow \tau \theta_i + (1 - \tau)\theta_i'$ and  $\phi' \leftarrow \tau\phi + (1 - \tau)\phi'$ for some learning rate $\tau>0$;
    \ENDIF
    \ENDFOR
\end{algorithmic}
\end{algorithm*}

\subsection{Post-Training Fitting of LQR}
\label{Sec:Optimization Formulation}
The third stage of our algorithm is to fit the output of the actor from Algorithm~\ref{Alg:TD3training} to a known stable LQR controller. As stated in the previous section, to fit the actor to an LQR controller, we just need to adjust the last layer weights $W_n$ and bias $b_n$ in \eqref{Eq:eff_weight} and \eqref{Eq:eff_bias}. We construct a least squares problem using  $W_n$ and $b_n$. Since we are only modifying the last layer of the actor, the features in the hidden layers are preserved\footnote{
The image processing community has researched the utilization of certain layers for tuning while maintaining hidden features in others \cite{sharif2014cnn, razavian2016visual, babenko2014neural, gong2014multi, yue2015exploiting}.
}. These optimization variables will be denoted as $W_o$ and $b_o$, respectively. Also any additional variables will have the subscript $o$. The desired parameters generally are the original $W_n$ and $b_n$ of the policy trained in simulation, staying close to the original values. Symbols without any of the subscripts mentioned will be taken as constants. The optimization problem modifies the last layer to fit the LQR optimal gain, $K$, from \eqref{Eq:lqr_control} in the linear region:
\begin{align}
\vspace{-1cm}
    \text{minimize} &\;\lVert W_{o}-W_{n} \rVert+\lVert b_{o}-b_{n} \rVert\nonumber\\
    \text{subject to} &\;W_{o}\hat{W}_{n-1}^{(m)} + K = 0, W_{o}\hat{b}_{n-1}^{(m)}+b_{o} = 0
\label{Eq:opt_1}
\vspace{-0.8cm}
\end{align}
\eqref{Eq:opt_1} can be relaxed by making the constraint on $K$ to be a cost in the objective. To ensure that the controller is at least stabilizing we check the real parts of the eigenvalues of the closed loop system. If they are stable we are done; if not, the weighting, $\nu_k$, increases before solving again as described by Algorithm \ref{Alg:optimize}. The second constraint can also be loosened, giving the following:
\begin{equation}
\begin{split}
    \text{minimize} &\;\lVert W_{o}-W_{n} \rVert+\lVert b_{o}-b_{n} \rVert+\nu_k\lVert W_{o}\hat{W}_{n-1}^{(m)} + K \rVert\\
    \text{subject to} &\;|W_{o}\hat{b}_{n-1}^{(m)}+b_{o}| \le \epsilon
\end{split}
\label{Eq:opt_2}
\end{equation}
The stability check is done by ensuring the real part of the maximum eigenvalue of the closed loop system, i.e., plugging \eqref{Eq:lqr_control} into \eqref{Eq:linear_sys}, is negative, giving the following:
\begin{equation}
\mathcal{R}eal(\lambda_{max} (A + B(W_{o}\hat{W}_{n-1}^{(m)} + W_{o}\hat{b}_{n-1}^{(m)}+b_{o}))) < 0
\label{Eq:closedloop_stable}
\end{equation}

\subsection{Regularization \& Linearity Maintenance}
\label{Sec:regularization}
Lastly, we state some potential issues of the framework and solutions to solve those issues.
By \eqref{Eq:eff_weight} and \eqref{Eq:eff_bias} the weights and biases from the previous layers can cause the linear region to collapse, and the fit to LQR from Sec. \ref{Sec:Optimization Formulation} can potentially  change the policy in the nonlinear region. As a countermeasure to these adverse possibilities, we propose adding regularization terms to the loss when updating the policy. To maintain the linear region during training, the gain on input $x$ from layer to layer should be kept relatively close to unity to keep the input from expanding past the desired linear region or saturating, and the effects of previous layers need to be kept from cancelling out $b_l$ at each individual layer. 

The two-layer example in Sec. \ref{Sec:Linear Region of Neural Networks} can be used again to demonstrate how to design regularization terms. To have a one-to-one scaling with the input, the gains at every layer need to be adjusted accordingly. We do this by penalizing the weights at every layer towards a scaled version of itself. This is achieved by adding weight regularizers to the loss before updating the policy. The $1^{\text{st}}$ and $2^{\text{nd}}$ layers regularization terms are respectively: $\lVert W_1 - \text{sign}(W_1)\alpha \rVert$ and $\lVert W_2W_{1}^{(m)} - \text{sign}(W_2W_{1}^{(m)})\alpha \rVert$ with row scaling, $\alpha > 0$. 

At every layer there is a hard restriction that does not allow the biases to be updated during training if $|b_{l,j}|<m$. With this in mind, the first layer will always have a preserved linear region with no additional loss term needed. However, simply restricting the magnitude of $b_2$ does not guarantee $|\hat{b}_{2,j}|<m$. Thus, the additional biases that filter from the preceding layer are penalized and driven towards zero by adding $\lVert W_2b_{1}^{(m)} \rVert$ to the loss. Combining this with the previous weight scaling loss terms, we can generate the general loss terms for every layer:
\begin{equation}
J = \lVert W_n\hat{W}_{n-1}^{(m)} - \text{sign}(W_n\hat{W}_{n-1}^{(m)})\alpha \rVert + \lVert W_n\hat{b}_{n-1}^{(m)} \rVert
\label{Eq:maintain_linear}
\end{equation}

Finally, two additional regularization terms are added to keep the policy closer to what we want to fit. Doing this during training has the added benefit of helping preserve the nonlinear regions of the policy when fitting the last layer $W_n$ and $b_n$ to output the LQR. An unintentional side affect seems to be that this regularizer acts like a supervisor near the linear region (Sec. \ref{Sec:Results} for more details).
\begin{algorithm}[b!]
   \caption{Optimize Actor Last Layer Parameters}
   \label{Alg:optimize}
\begin{algorithmic}[1]
    \STATE \textbf{Input:} choose exact fit (Eq. \ref{Eq:opt_1}) or relaxed (Eq. \ref{Eq:opt_2})
    \IF{exact fit}
    \STATE Solve Eq. \ref{Eq:opt_1} for last layer $\phi_{n}:=(W,b)_{\phi_{n}}$
    \STATE $\phi_n\leftarrow (W_o, b_o)$
    \ELSE
    \WHILE{not closed loop stable (Eq. \ref{Eq:closedloop_stable})}
    \STATE Solve Eq. \ref{Eq:opt_2} for last layer $\phi_{n}$
    \STATE $\phi_n \leftarrow (W_o, b_o)$
    \STATE Increment $\nu_{K}$
    \ENDWHILE
    \ENDIF
\end{algorithmic}
\end{algorithm}
 \eqref{Eq:opt_1} constraints can be formulated into a loss to penalize the policy for being far away from the desired controller, respectively as:
\begin{align}
    &\lVert W_o\hat{W}_{n-1}^{(m)} + K \rVert \label{Eq:maintain_k}\\
    &\;\lVert W_o\hat{b}_{n-1}^{(m)}+b_o \rVert \label{Eq:maintain_bz}
\end{align}
The total regularized loss for the actor, $\text{Reg}((W,b)_{\phi})$, is the sum of \eqref{Eq:maintain_linear} at every layer with \eqref{Eq:maintain_k} and \eqref{Eq:maintain_bz}.

%% file: Sections/6Results.tex
\section{Results}
\label{Sec:Results}
\subsection{Architecture and Computation}
The ``bias-shifted'' network architecture uses the original TD3 implementation as a template as stated in Section~\ref{sec:alg}. The actor network consists of 3 layers: linear $s\text{-}dim \times 512$ with ReLU activation, linear $512 \times 256$ with ReLU activation, and linear $256 \times a\text{-}dim$ with hyperbolic tangent activation, where $dim$ stands for dimension. During training dropout was used at all layers of the actor. Meanwhile, the critics' networks are: linear $s\text{-}dim \times 512$ with ReLU activation, linear $512 \times 256$ with Swish activation, and linear $256 \times a\text{-}dim$ with no activation.

\subsection{PyBullet Simulation}
To show that linear region preserving modifications still allow the network to remain expressive, the bias shifted network was trained alongside other TD3 variations using the OpenAI Gym library and PyBullet environments. A comparison of the performance in the HalfCheetah environment can be seen in Fig. \ref{Fig:bias_no_bias} while other environments can be found in Appendix \ref{Sec:other_envs}. Fig. \ref{Fig:bias_no_bias} shows the individual pieces of the ``bias-shifted'' network implemented in TD3, as well as the full implementation, did not cause a loss of expressiveness. Even despite the restriction on bias updates, all variations performed on par with the original TD3 implementation, demonstrating the networks can still adapt to complex tasks.

\begin{figure}[ht!]
\centerline{\includegraphics[width=0.8\columnwidth]{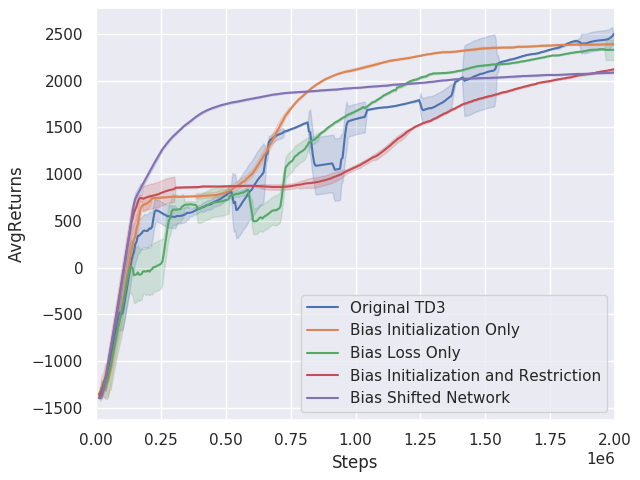}}
\caption{Performance comparison of TD3 variations and the bias shifted model in the HalfCheetah PyBullet Environment. Ten evaluations were used with max episode length of 10000 steps. \emph{Original TD3} (\emph{blue}): the original TD3 implementation with a reset condition added; otherwise, it could not converge at all. \emph{Bias Initialization Only} (\emph{orange}): TD3 with bias initialization and our architecture. \emph{Bias Loss Only} (\emph{green}): TD3 with added regularization terms. \emph{Bias Initialization and Restriction} (\emph{red}): TD3 with our actor-critic with initialization and bias magnitude restrictions. \emph{Bias Shifted Network} (\emph{purple}): TD3 with the full ``bias-shifted'' implementation.}
\label{Fig:bias_no_bias}
\end{figure}

\subsection{Swing-up Pendulum}
To illustrate the advantages of the ``bias-shifted'' network, the swing-up pendulum was chosen, as it is a nonlinear system that can be linearly approximated at the top near its balancing point. In addition, it is widely studied and used to model many real world tasks such as legged robotic locomotion \cite{raibert1984experiments, kajita20013d}. The pendulum has its torque limited such that it cannot balance at the top for initial conditions outside a certain region. Thus, starting at the bottom, the policy must use a ``pump up'' motion by gathering energy on successive swings to reach the top.

A pendulum was constructed from a custom brushless DC motor, a carbon fiber tube, and an aluminum mass. In addition, a custom pendulum simulation environment was created for training with the same measured physical parameters (mass = $0.4$ kg, length = $0.37$ m, and damping = $0.1$ Ns/m). The maximum allowed torque was set to 0.8 Nm, about half the required torque needed to swing the pendulum to the top from any starting position. Finally, the reward function follows the LQR cost from \eqref{Eq:lqr_cost_int} plus an exit reward of $1000.0$ with the following weighting matrices:
\begin{equation}
    Q = \begin{bmatrix}
    1.0 & 0 \\
    0 & 0.1 \\
    \end{bmatrix} \qquad
    R = 0.001
\label{eq:pendulum_cost_weighting}
\end{equation}
The mass on the real pendulum was placed off center of the pendulum arm to add modeling discrepancies. Additional nonlinearities such as friction, backlash, and the motor feedback controller were also purposefully unmodeled. An LQR controller was tuned on the hardware pendulum for use during training and optimization.

\begin{figure}[ht!]
\centerline{\includegraphics[width=0.8\columnwidth]{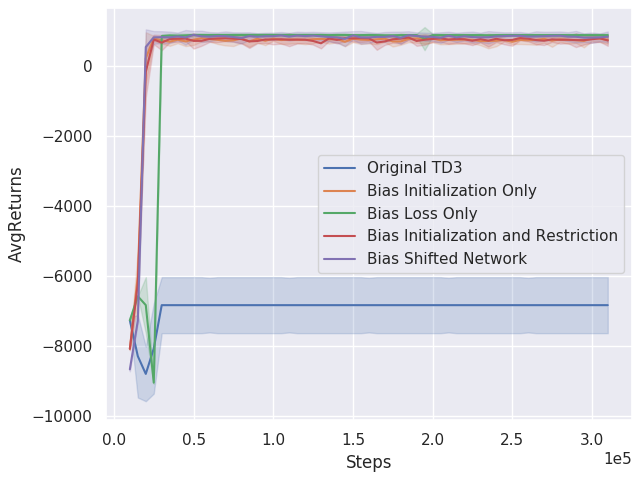}}
\caption{Performance comparison between TD3 variations and TD3 with our ``bias-shifted'' methods for a custom swing-up pendulum using the forward Euler method to propagate its dynamics. Ten evaluations were used with a max episode length of 1000 time steps. Please refer to the caption in Fig. \ref{Fig:bias_no_bias} for legend definitions.}
\label{Fig:swing_pendulum_comparison}
\end{figure}

\begin{figure}[ht]
\centerline{\includegraphics[width=\columnwidth]{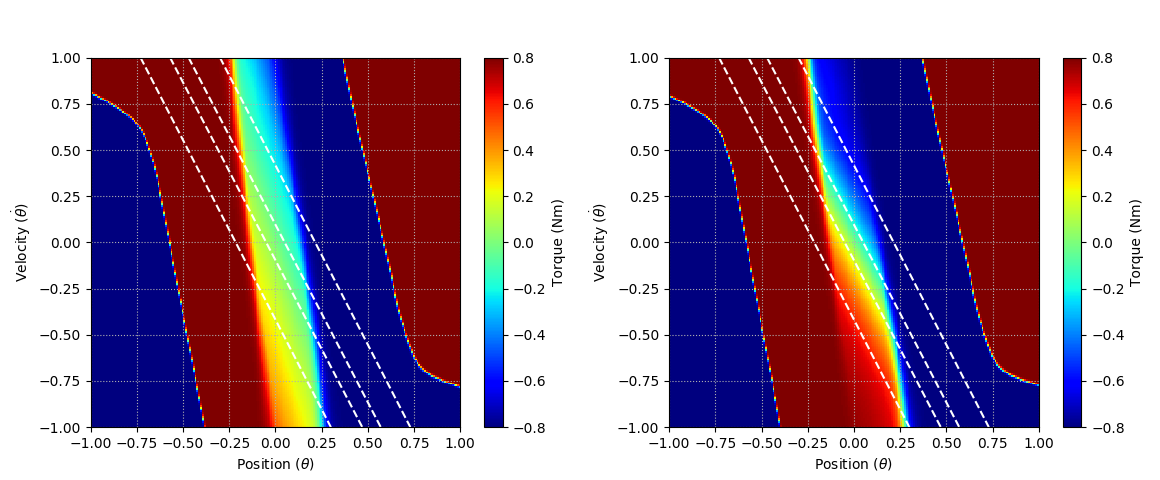}}
\caption{A policy was trained in simulation and sampled at different points in the state space. \emph{Red} corresponds to saturation at the positive torque limit while \emph{blue} corresponds to saturation at the negative torque limit. All hues in between reflect intermediate torques from $[-0.8, 0.8]$. The white dotted lines represent contours of the LQR policy from Fig.~\ref{fig:pendulum}. \emph{Left}: The bias shifted policy after training on the swing up pendulum environment for 300k samples. \emph{Right}: The bias shifted policy after the optimization was performed to fit the hardware tuned LQR controller. As can be seen, the policy after the optimization is transformed to match the output of the LQR controller in a region around the origin.}
\label{Fig:before_and_after}
\end{figure}

The TD3 variations were trained on simulated swing-up pendulum environment as shown in Fig.~\ref{Fig:swing_pendulum_comparison}. As can be seen by the original TD3 performance, the parameters for this environment yield a difficult problem to solve where the original TD3 cannot even converge under 300k samples. We suspect that in creating a linear region and placing regularizing terms to fit the LQR controller, the ``bias-shifted'' network and its counterparts are subjected to a form of supervised learning in which the stabilizing region is an example to imitate. Thus, the policy only needs to learn how to generate enough speed such that the controller can attract the pendulum to the top.
A ``bias-shifted'' policy trained on 300k samples in the simulated pendulum environment before optimization is shown in Fig. \ref{Fig:before_and_after}. As can be seen, the swing up policy has a linear feedback region near the origin but nonlinear control further away. Meanwhile, the optimized policy fit to the tuned LQR controller is also depicted in Fig. \ref{Fig:before_and_after}. While most of the policy remains untouched, the region near the origin is transformed into the LQR from Fig.~\ref{fig:pendulum}.

To verify the region of attraction on hardware, a bias shifted network was trained on the swing-up simulation environment for 100k samples. As a comparison, a policy with the original TD3 implementation and one with TD3 and ADR were also trained. The bias shifted network and the TD3 implementation performed similarly to that in Fig. \ref{Fig:swing_pendulum_comparison}. TD3 alone was never able to learn the swing up or stabilizing behavior. Meanwhile, TD3 with ADR was started at a max torque of $\pm1.0$ (an easier task) and converged to use a max torque of $\pm0.8$ after 80k samples.  
\begin{table}[h]
\caption{Cost comparison on the hardware pendulum with weighting matrices from \eqref{eq:pendulum_cost_weighting} and $x_0\in(-0.4, 0.4)$ rads for $5\,$s. Each policy was evaluated at 50k sampling intervals with a random initialization. The 100k version of TD3 was not run since the policy did not converge and yielded dangerous velocities of 10 rad/s.}
\label{tab:hardware_results}
\vskip 0.02in
\begin{small}
\begin{sc}
\begin{tabular}{lcccr}
\toprule
Samples& 0k & 50k & 100k \\
\midrule
TD3    & -3030 $\pm$ 90 & -3060 $\pm$ 1070 & - \\
TD3 (ADR) & -3090 $\pm$ 100 & -1440 $\pm$ 860 & -8.8 $\pm$ 2.8 \\
Bias Shift & -7.8 $\pm$ 2.7 & -7.9 $\pm$ 2.3 & -8.6 $\pm$ 3.5 \\
\bottomrule
\end{tabular}
\end{sc}
\end{small}
\vskip -0.01in
\end{table}
At different checkpoints (0, 50k, and 100k total samples trained), the ``bias-shifted'' and TD3 implementations were evaluated on hardware. Each policy underwent 10 trials with a starting position sampled uniformly from $(-0.4, 0.4)$ rads with $\dot{x}_0=0$. The policy was then ran for 5s and the total cost was reported, as seen in Table \ref{tab:hardware_results}. The TD3 implementation was not run at 100k samples because the policy never converged and continually spun the pendulum, yielding a dangerous max velocity of 10 rad/s. As can be seen, the ``bias-shifted'' network can be optimized after any given number of samples (even at zero training samples), to yield a stabilizing region. Meanwhile, ADR must reach a certain number of samples in order to reach the same level stability.

\subsection{Computer Hardware/Computation}
Fitting the LQR controller can be done at any time during training to create a region of attraction. Even the most expensive training set for the bias shifted network on the pendulum (300k samples) was equivalent to only 8.5 real world hours of experience. Due to this low cost of computation, training was done on a single machine with an Intel i7-7700X processor and Nvidia 1080 Ti graphics card.

%% file: Sections/7Conclusion.tex
\section{Conclusion}
In this paper, ``bias-shifted'' networks were introduced to allow an linear region of the state space to be passed through. To prevent the region from collapsing while maintaining most of the policy's nonlinearities, bias update restrictions and regularization terms were added. Consequently, prior knowledge about the system can presumably be embedded directly into the policy during training. An optimization approach was also introduced that allowed tuning of last layer parameters to match an LQR controller, known \emph{a priori} to be stable, as the output in the linear region. The approach outperformed TD3 with DR and the original TD3 implementation for a simulated swing-up pendulum. For practical hardware applications of deep reinforcement learning, ``bias-shifted'' networks provide an alternative approach with guarantees when collecting data from hardware.

%% file: Sections/A1TD3Algorithm.tex
\section{TD3 Original Algorithm}
\label{Sec:original_td3}
Here we show the pseudocode for the unaltered TD3 policy gradient algorithm from \cite{fujimoto2018addressing}. Since our method only involves initialization, regularization terms in the loss, and a post-training update to the last layer weights, the overall structure of the original TD3 algorithm is maintained. Hence, its similarity to Algorithm~\ref{Alg:TD3training}.
\begin{algorithm}[htb!]
\caption{TD3 Original}
\label{Alg:TD3Original}
\begin{algorithmic}[1]
\STATE \textbf{Input:}\\ 
\STATE Number of iterations $T>0$; 
\STATE Number of iterations before updating targets: $d>0$;
\STATE Standard deviation used for exploration $\sigma_e>0$;
\STATE Standard deviation used for smoothing $\sigma_s>0$;
\STATE Policy smoothing range $(-c, c)$, $c>0$;
\STATE Polyak averaging $\tau>0$;
\STATE \textbf{Initialization:}
\STATE Initialize critic and actor networks: $Q_{\theta_1}$, $Q_{\theta_2}$ and $\pi_\phi$;
\STATE Initialize target networks: $\theta_1' \leftarrow \theta_1$, $\theta_2' \leftarrow \theta_2$, $\phi'\leftarrow \phi$
\STATE Initialize replay buffer with random samples $\mathcal{B}$;
\FOR{$t=1$ {\bfseries to} $T$}
\STATE Let $a \gets \pi_\phi(s) + \epsilon$, where $\epsilon \sim \mathcal{N}(0, \sigma_e)$;
\STATE Execute $a$ in state $s$ to get next state $s'$ and reward $r$;
\STATE Store ($s, a, r, s'$) in $\mathcal{B}$;
\STATE Sample $N$ transitions $\{s_k, a_k, r_k, s_k'\}_{k=1}^{N}$ from $\mathcal{B}$;
\STATE For each $k\in[N]$: let $\tilde a_k \leftarrow \pi_\phi'(s'_k) + \epsilon$, \\
$\;$ where $\epsilon \sim \mathrm{clip}(\mathcal{N}(0, \sigma_s), -c, c)$;
\STATE $y_k \leftarrow r_k + \gamma  \min_{i=1,2} Q_{\theta_i'}(s_k', \tilde a_k)$, \\
$\;$ where $\gamma\in(0,1)$ is a discount factor;
\STATE Update critics using least square fitting:\\
$\theta_i \gets \underset{\theta_i}{\text{argmin}}\: N^{-1}\sum_{k=1}^N(y_k-Q_{\theta_i}(s_k, a_k))^2$
\IF{$t$ mod $d$}
\STATE Update $\phi$ by policy gradient:\\
$\nabla_\phi J(\phi) \gets \frac{1}{N}\sum\nabla_a Q_{\theta_1}(s, a)|_{a=\pi_\phi}\nabla_\phi\pi_\phi(s)$
\STATE Update target networks:\\
$\theta_i' \leftarrow \tau \theta_i + (1 - \tau)\theta_i'$ and  $\phi' \leftarrow \tau\phi + (1 - \tau)\phi'$;
\ENDIF

\ENDFOR
\end{algorithmic}
\end{algorithm}

%% file: Sections/A2AllTrainedGraphs.tex
\section{Performance on Different Environments}
\label{Sec:other_envs}
\begin{figure}[!h]
	\centering
		\includegraphics[scale=0.35]{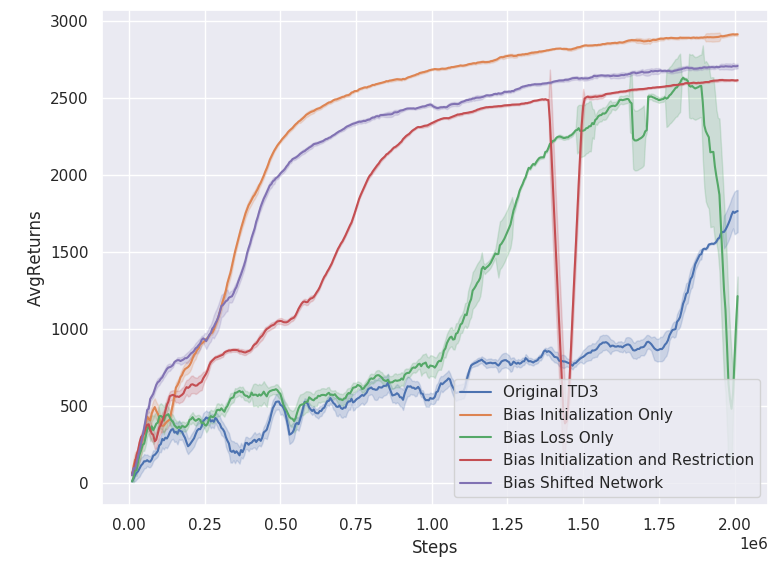}
	~~
		\includegraphics[scale=0.35]{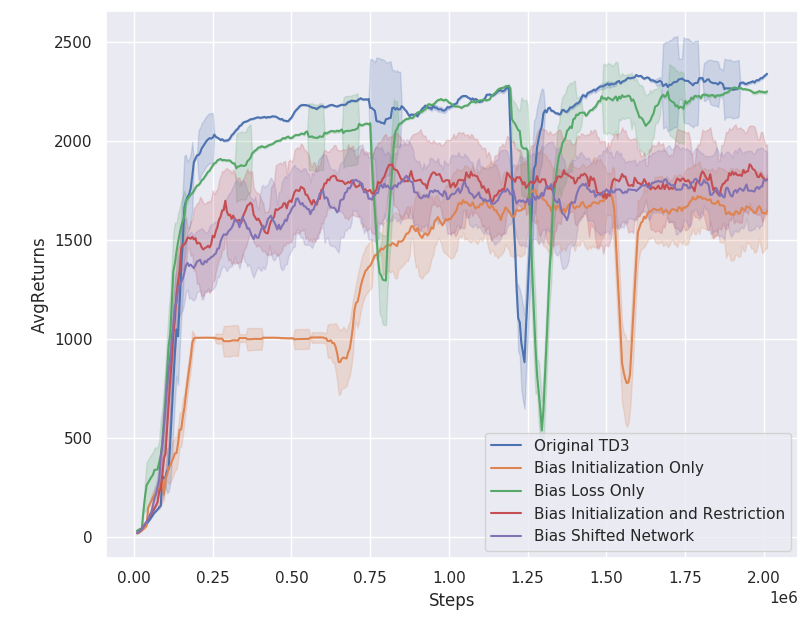}
	~~
    	\includegraphics[scale=0.35]{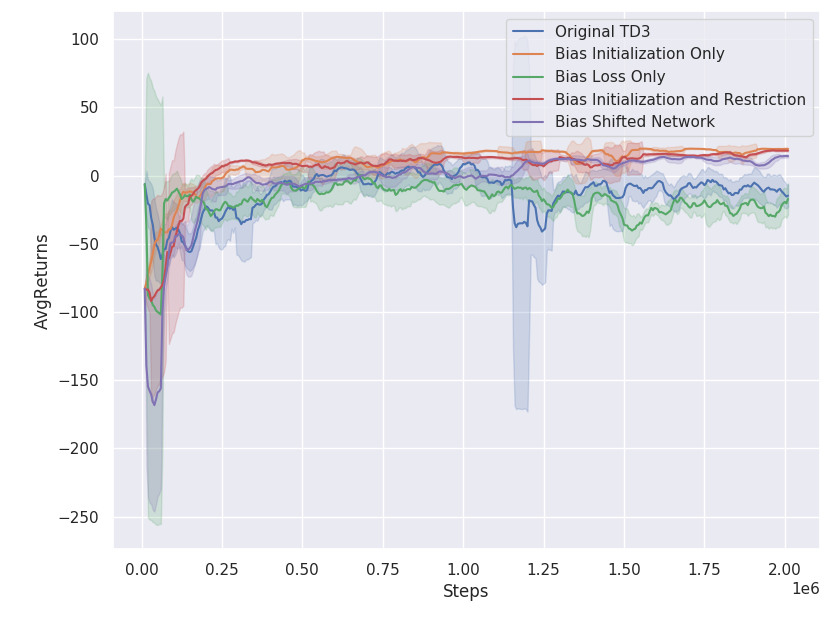}
	\caption{Performance comparison of TD3 variations and the bias shifted model in multiple PyBullet environments: \emph{Ant} (\emph{top}), \emph{Hopper} (\emph{middle}), and \emph{Reacher} (\emph{bottom}). Ten evaluations were used with max episode length of 10000 steps. \emph{Original TD3 (blue)}: the original TD3 implementation with a reset condition added; otherwise, it could not converge at all. \emph{Bias Initialization Only (orange)}: TD3 with bias initialization and our architecture. \emph{Bias Loss Only (green)}: TD3 with added regularization terms. \emph{Bias Initialization and Restriction (red)}: TD3 with our actor-critic with initialization and bias magnitude restrictions. \emph{Bias Shifted Network (purple)}: TD3 with the full bias shifted implementation.}
	\label{Fig:pybullet_sims}
\end{figure}

In this section we observe the expressiveness of our methods compared to that of the original TD3 algorithm for other OpenAI PyBullet environments. In Fig.~\ref{Fig:pybullet_sims} the plots from top to bottom are of the \emph{Ant}, \emph{Hopper}, and \emph{Reacher} environments. The original TD3 algorithm is shown in blue. The other colors correspond to certain components of our ``bias-shifted'' method with TD3. See the caption of Fig.~\ref{Fig:pybullet_sims} for more details.

We also noticed that the original TD3 did not perform consistently in these environments. Regardless, the main take away from these graphs is that our methods are not detrimental to the expressiveness of the actor.

%% file: Sections/A3PolicyEvolution.tex
\section{Visualization of Policy Outputs}
In this section we visualize the post-training fitting to the LQR for our full ``bias-shifted'' implementation for the results shown in Table~\ref{tab:hardware_results}. The white lines in the figures trace the original LQR optimal policy seen in Fig.~\ref{fig:pendulum}.

\begin{figure}[ht]
\centerline{\includegraphics[width=\columnwidth]{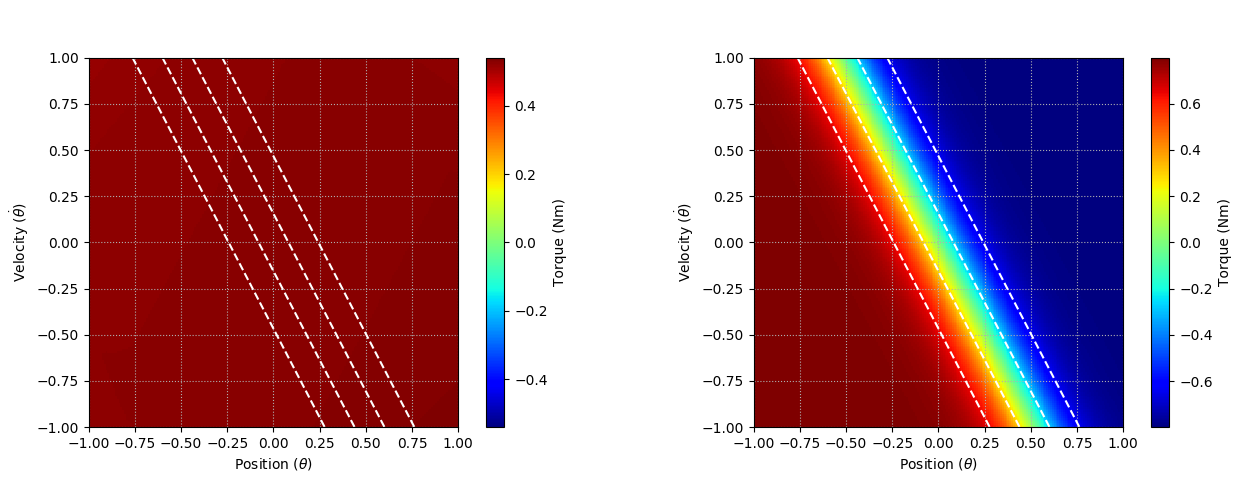}}
\caption{Comparison with 0K samples of control policy before optimization \textit{(left)} and after optimization \textit{(right)} over the state space. The control ranges from [-0.8, 0.8] Nm torque. The white dashed lines show the level sets of the LQR controller.}
\label{Fig:pybullet sims}
\end{figure}

\begin{figure}[ht]
\centerline{\includegraphics[width=\columnwidth]{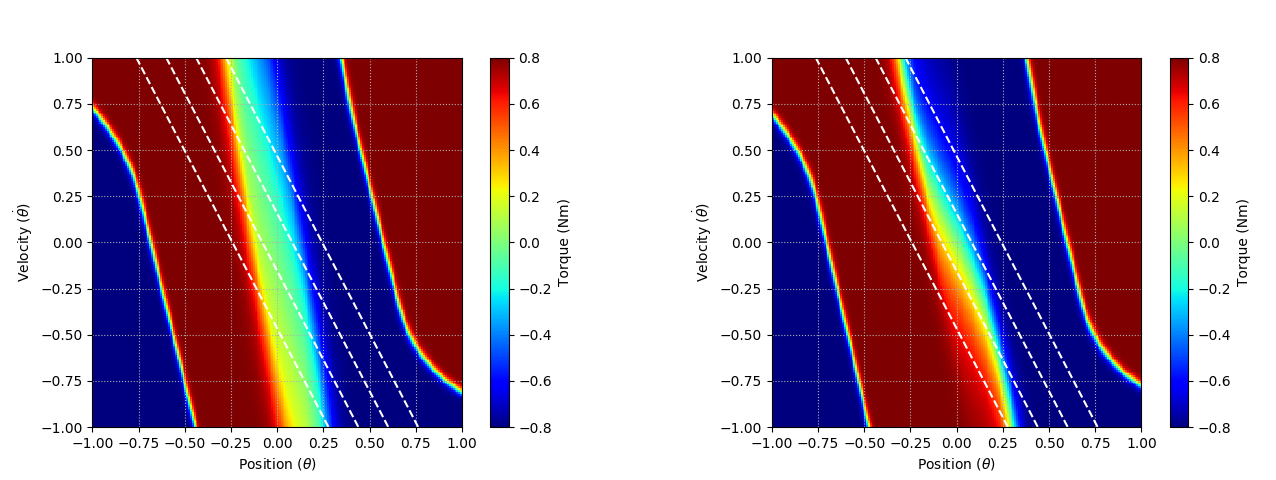}}
\caption{Comparison with 50K samples of control policy before optimization \textit{(left)} and after optimization \textit{(right)} over the state space. The control ranges from [-0.8, 0.8] Nm torque. The white dashed lines show the level sets of the LQR controller.}
\label{Fig:pybullet sims}
\end{figure}

\begin{figure}[ht]
\centerline{\includegraphics[width=\columnwidth]{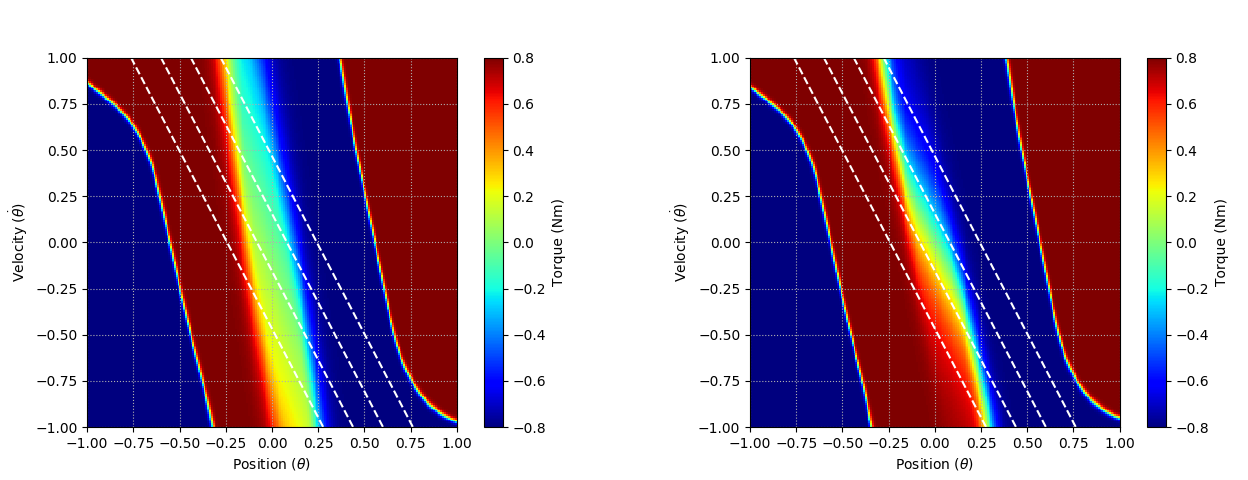}}
\caption{Comparison with 100K samples of control policy before optimization \textit{(left)} and after optimization \textit{(right)} over the state space. The control ranges from [-0.8, 0.8] Nm torque. The white dashed lines show the level sets of the LQR controller.}
\label{Fig:pybullet sims}
\end{figure}

\begin{figure}[ht]
\centerline{\includegraphics[width=\columnwidth]{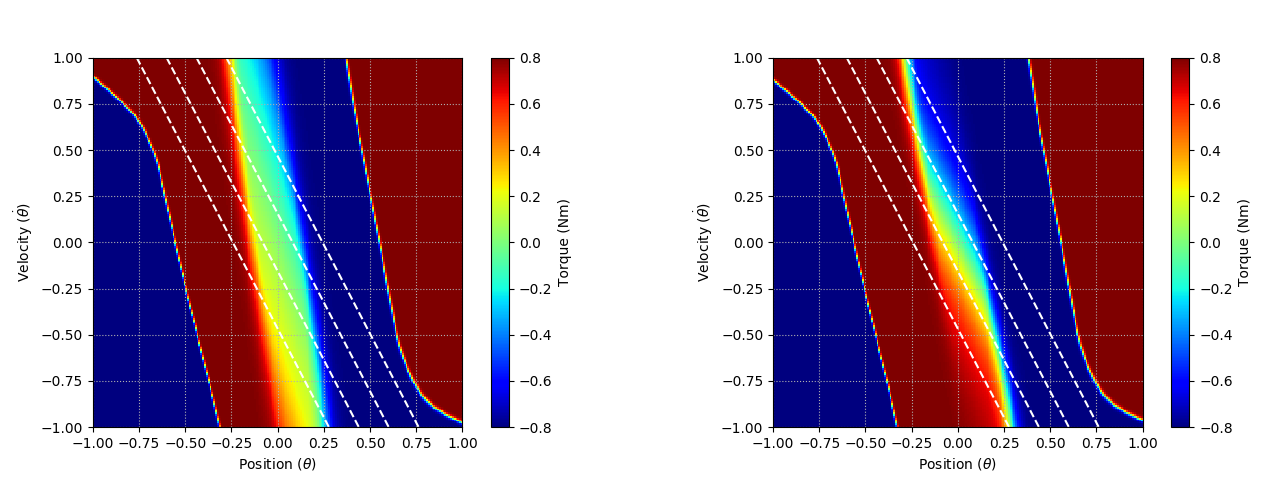}}
\caption{Comparison with 150K samples of control policy before optimization \textit{(left)} and after optimization \textit{(right)} over the state space. The control ranges from [-0.8, 0.8] Nm torque. The white dashed lines show the level sets of the LQR controller.}
\label{Fig:pybullet sims}
\end{figure}

\begin{figure}[ht]
\centerline{\includegraphics[width=\columnwidth]{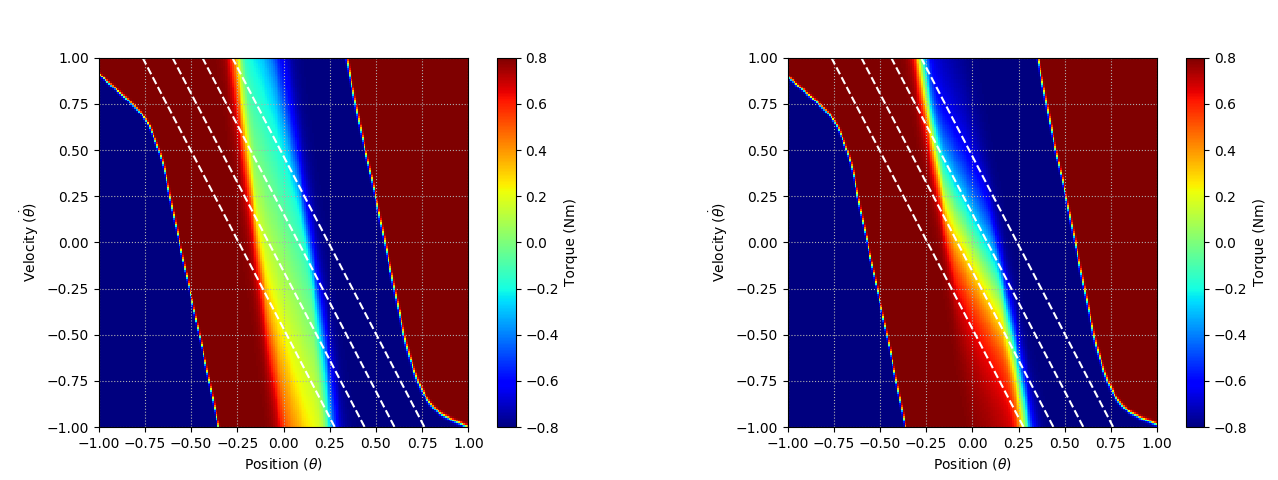}}
\caption{Comparison with 200K samples of control policy before optimization \textit{(left)} and after optimization \textit{(right)} over the state space. The control ranges from [-0.8, 0.8] Nm torque.}
\label{Fig:pybullet sims}
\end{figure}

\begin{figure}[ht]
\centerline{\includegraphics[width=\columnwidth]{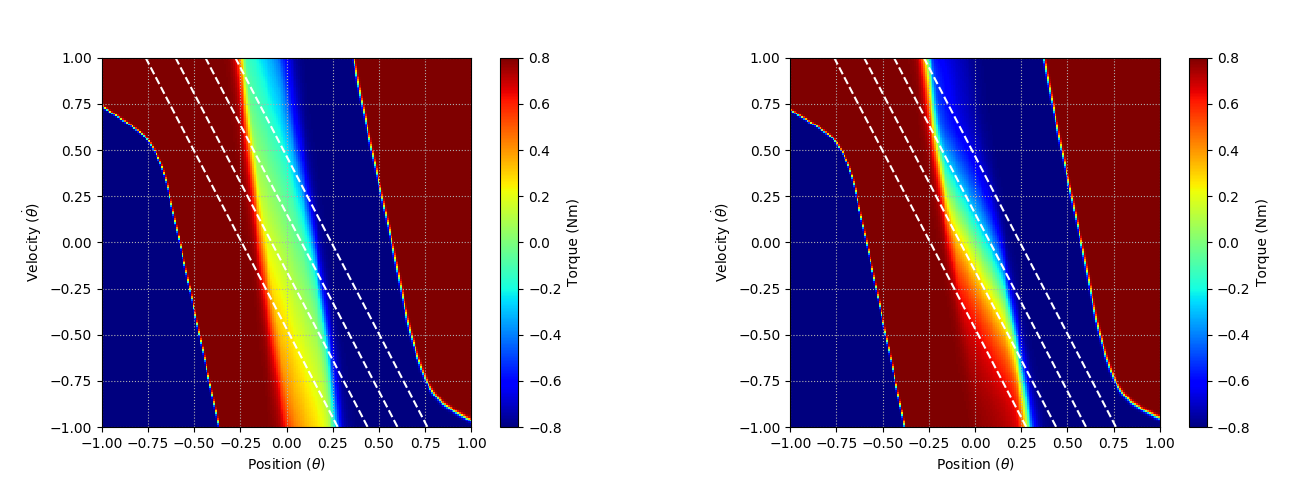}}
\caption{A 250K sample comparison of control policy before optimization \textit{(left)} and after optimization \textit{(right)} over the state space. The control ranges from [-0.8, 0.8] Nm torque. The white dashed lines show the level sets of the LQR controller.}
\label{Fig:pybullet sims}
\end{figure}
\vfill